\documentclass[10pt,twocolumn,letterpaper]{article}

\usepackage{cvpr}              % To produce the CAMERA-READY version
\usepackage[dvipsnames]{xcolor}

\usepackage{adjustbox}

\definecolor{cvprblue}{rgb}{0.21,0.49,0.74}
\usepackage[pagebackref,breaklinks,colorlinks,citecolor=cvprblue]{hyperref}

\def\paperID{9688} % *** Enter the Paper ID here
\def\confName{CVPR}
\def\confYear{2024}

\title{Pre-trained Model Guided Fine-Tuning for Zero-Shot Adversarial Robustness}
\author{
    Sibo Wang\textsuperscript{1,2} \quad 
    Jie Zhang\textsuperscript{1,2} \quad  
    Zheng Yuan\textsuperscript{1,2} \quad  
    Shiguang Shan\textsuperscript{1,2}\\
    \textsuperscript{1}Institute of Computing Technology, Chinese Academy of Sciences, Beijing, China\\
    \textsuperscript{2}University of Chinese Academy of Sciences, Beijing, China\\
    {\tt\small \{wangsibo22z, zhangjie, sgshan\}@ict.ac.cn , zheng.yuan@vipl.ict.ac.cn}
}

\begin{document}
\captionsetup[table]{skip=4pt} % 调整为你想要的距离
\maketitle
\begin{abstract}
Large-scale pre-trained vision-language models like CLIP have demonstrated impressive performance across various tasks, and exhibit remarkable zero-shot generalization capability, while they are also vulnerable to imperceptible adversarial examples. 
Existing works typically employ adversarial training (fine-tuning) as a defense method against adversarial examples. 
However, direct application to the CLIP model may result in overfitting, compromising the model's capacity for generalization.
In this paper, we propose Pre-trained Model Guided Adversarial Fine-Tuning (PMG-AFT) method, which leverages supervision from the original pre-trained model by carefully designing an auxiliary branch, to enhance the model's zero-shot adversarial robustness.
Specifically, PMG-AFT minimizes the distance between the features of adversarial examples in the target model and those in the pre-trained model, aiming to preserve the generalization features already captured by the pre-trained model.
Extensive Experiments on 15 zero-shot datasets demonstrate that PMG-AFT significantly outperforms the state-of-the-art method, improving the top-1 robust accuracy by an average of 4.99\%.
Furthermore, our approach consistently improves clean accuracy by an average of 8.72\%.
Our code is available at \href{https://github.com/serendipity1122/Pre-trained-Model-Guided-Fine-Tuning-for-Zero-Shot-Adversarial-Robustness}{here}.\footnote{https://github.com/serendipity1122/Pre-trained-Model-Guided-Fine-Tuning-for-Zero-Shot-Adversarial-Robustness}
\end{abstract}
\vspace{-3mm}
% title candidate
% 1 Leverage Knowledge from the Foundational Model for Zero-Shot Adversarial Robustness
% 2 Original Pre-trained Model Supervised(Guided) Adversarial Fine-Tuning for Zero-Shot Adversarial Robustness
% Pre-trained Model Guided Zero-Shot Adversarial Fine-Tuning

% Method name
% Pre-trained Model Guided Adversarial Fine-Tuning (OPMG FT)AT    
\section{Introduction}
\label{sec:intro}
\hspace*{0.422cm}Vision-language models pre-trained on large dataset have achieved impressive success in numerous tasks, such as image classification \cite{jia2021scaling,radford2021learning}, text-to-image generation \cite{reddy2021dall,ramesh2021zero}, image caption \cite{li2022blip,chen2022visualgpt}.
They have also showcased excellent zero-shot generalization ability.
As for CLIP \cite{radford2021learning}, given a test image and a set of candidate class labels, it computes the similarity between the image embedding and the embedding of each candidate class label and predicts the class as the one with the highest similarity.
\begin{figure}[t]
    \centering
    % 第一个子图
    \begin{subfigure}[t]{0.95\linewidth}
        \centering
        \includegraphics[height=0.65\linewidth, width=0.9\linewidth]{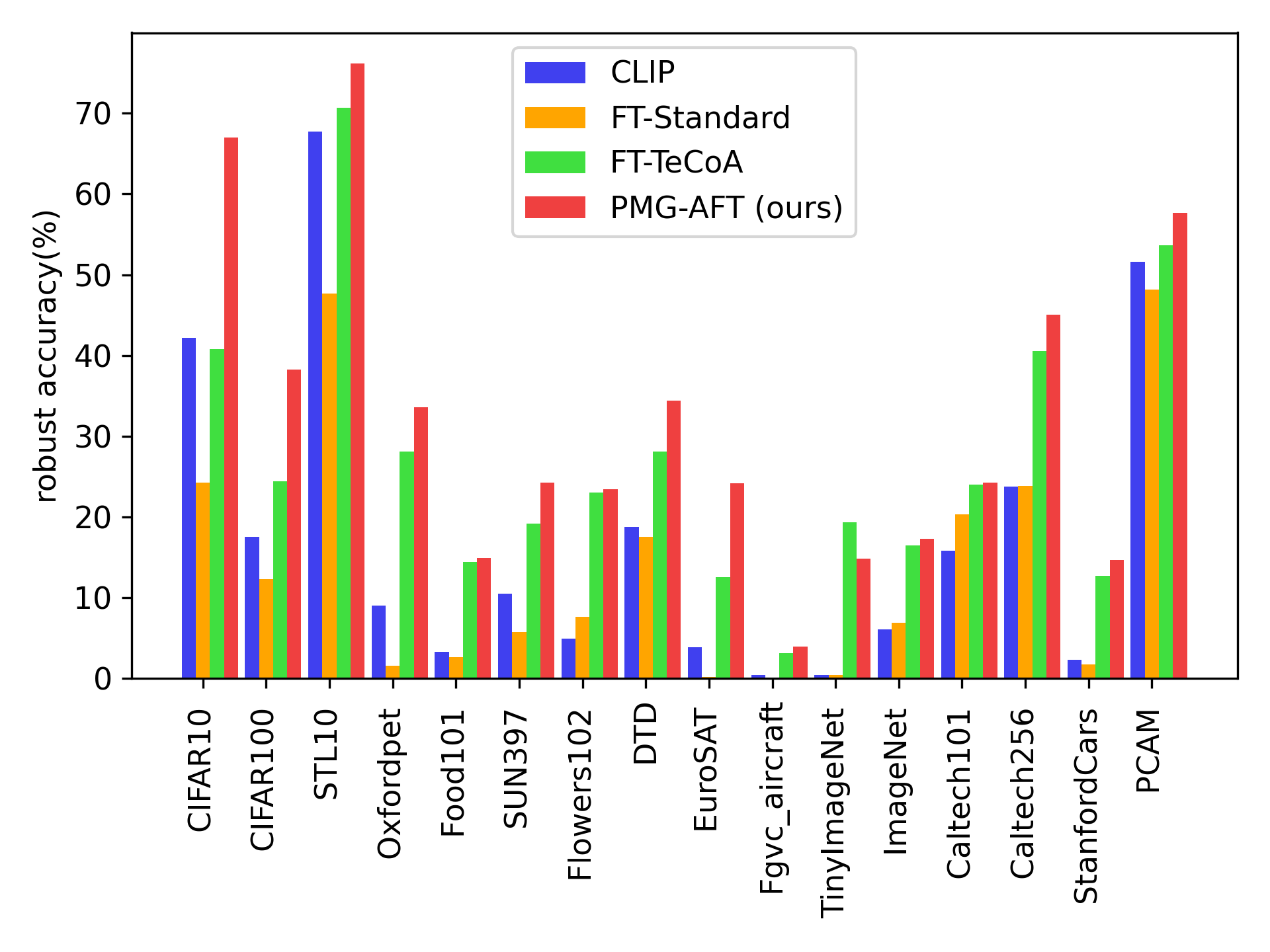}
        \vspace{-1mm}
        \caption{Robust Accuracy}
        \label{fig:fig1a}
    \end{subfigure}
    % \vspace{1cm} % 设置子图之间的垂直间距

    % 第二个子图
    \begin{subfigure}[t]{0.95\linewidth}
        \centering
        \includegraphics[height=0.65\linewidth, width=0.9\linewidth]{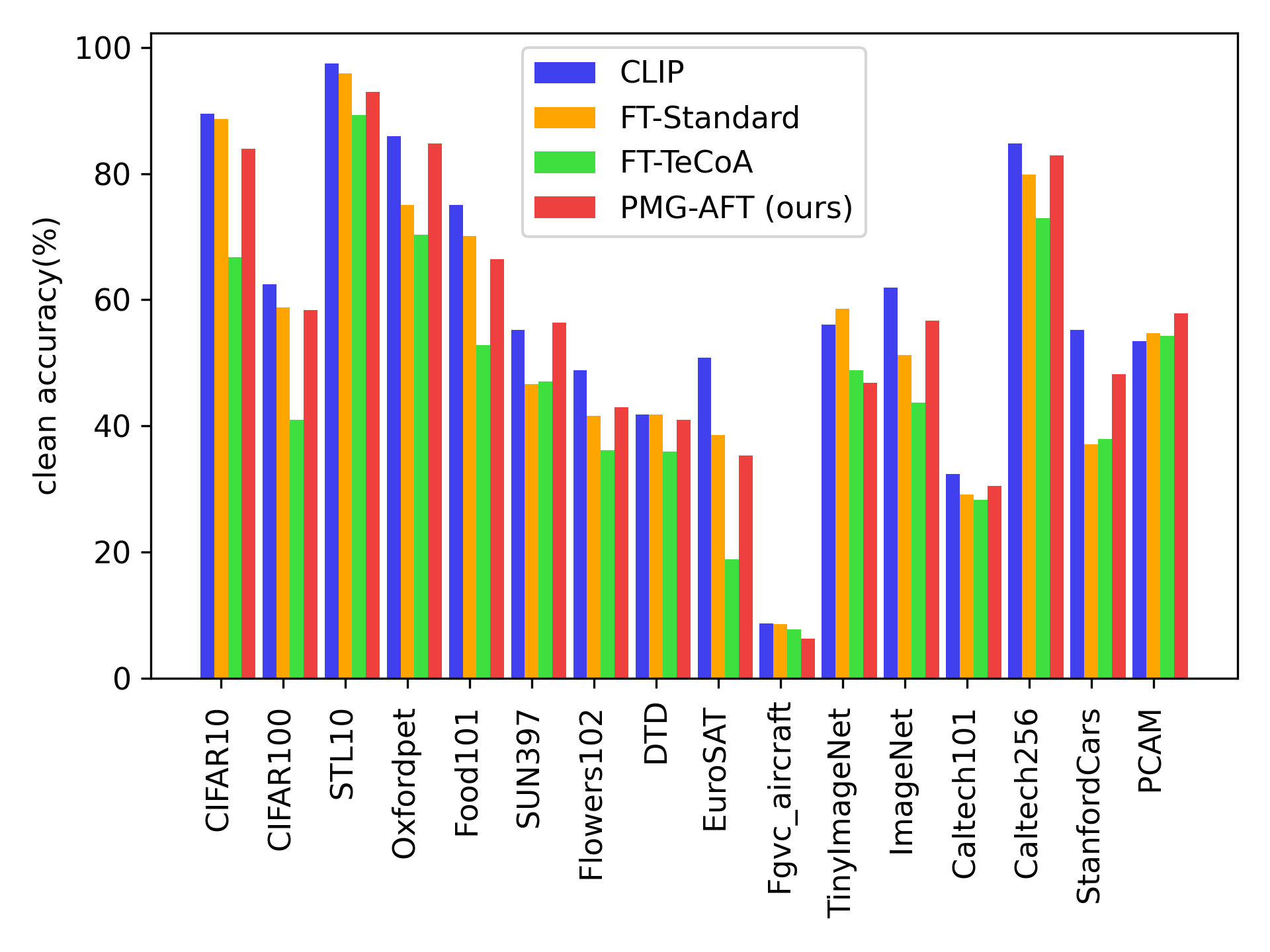}
        \vspace{-1mm}
        \caption{Clean Accuracy}
        \label{fig:fig1b}
        \vspace{-2mm}
    \end{subfigure}

    \caption{Zero-shot robust accuracy (a) and clean accuracy (b) of CLIP and CLIPs fine-tuned on TinyImageNet \cite{deng2009imagenet} using various methods across multiple datasets. FT-Standard: CLIP fine-tuned on clean samples. FT-TeCoA: CLIP fine-tuned using \cite{mao2022understanding}. PMG-AFT (ours): CLIP fine-tuned by our method.}
    %(a) 原始的CLIP模型以及在干净下游任务数据集(TinyImageNet)上微调在面对对抗样本时的表现很脆弱，通过TeCoA对抗微调可以在一定程度上提高模型对于下游未见任务的鲁棒性,我们的方法可以进一步提升这种鲁棒性 (b) TeCoA带来的鲁棒性的提升是以干净样本的准确率的大量降低作为代价的，我们的方法在面对干净样本时表现出与CLIP comparable的水平。
    \label{fig:figure1}
    \vspace{-6mm}
\end{figure}
While contemporary researches predominantly focus on enhancing their performance \cite{li2022blip}, comparatively less attention is devoted to investigating their robustness problem \cite{zhao2023evaluating}.
Deep models, including those at a large scale, are well-known to be vulnerable to adversarial examples \cite{goodfellow2014explaining,madry2017towards}.
By introducing deliberately designed imperceptible perturbations, it is easy to induce model failures.
Since an increasing number of large-scale models are deployed in security-related downstream tasks, it is imperative to enhance the robustness of such models.
To address the threat of adversarial examples, numerous defense methods have been proposed.
Among them, adversarial training \cite{madry2017towards,zhang2019theoretically} is considered one of the most effective defense strategies.
It incorporates adversarial examples into the training dataset during the training phase, significantly enhancing the robustness of deep neural networks.

% However, adversarial training primarily faces two significant challenges: on one hand, the effectiveness of the defense is closely tied to the type and strength of the attack, making it difficult to generalize to all attack scenarios; on the other hand, since it necessitates the concurrent generation of adversarial examples, it is computation-consuming.
% For large-scale models, fine-tuning is commonly employed to adapt to downstream tasks. This approach obviates the need to train a robust model from scratch, significantly reducing the resources required for adversarial training. 
% Meanwhile, during the pre-training phase, large-scale models acquire features with commendable generalization capabilities, which aid in defending against unseen types of attacks.
% Consequently, adversarial fine-tuning on large-scale models addresses the challenges of adversarial training from both these aspects in theory.
%这段写的有问题 问题应该是对各种下游任务的泛化性 不是对不同攻击的泛化性 但是对攻击能不能泛化我还没做实验

Undertaking adversarial training from scratch for large-scale models, however, may be an impractical approach.
Adversarial training is computation-consuming since it necessitates the concurrent generation of adversarial examples, especially in models with massive parameters. 
For large-scale models, fine-tuning is often employed to adapt to downstream tasks, significantly reducing the required computational resources. 
Therefore, applying adversarial training techniques to fine-tuning is an effective method to enhance the robustness of large models.
However, if we employ adversarial fine-tuning directly for large-scale models, there may be a tendency for the model to overfit the fine-tuning dataset \cite{mao2022understanding}. 
This overfitting leads to the model losing the superior generalization capabilities garnered during its pre-training phase.

A recent study \cite{mao2022understanding} explores a similar problem, defining it as zero-shot adversarial robustness. 
It investigates the zero-shot generalization ability of the CLIP model when dealing with adversarial examples.
From the perspective of adversarial examples generation, text supervision is introduced for generating adversarial examples, which are used for adversarial fine-tuning.
As illustrated in Fig. \ref{fig:fig1a}, the method with text supervision, \ie, FT-TeCoA \cite{mao2022understanding}, indeed improves the zero-shot robust accuracy across numerous unseen datasets, compared with both the original CLIP and the CLIP fine-tuned on clean datasets, which we refer to as FT-Standard.
However, this improvement is not sufficient and comes at the cost of a substantial decrease in the accuracy of clean samples, as shown in Fig. \ref{fig:fig1b}.
This suggests that adversarial fine-tuning still faces challenges in improving the model's adversarial robustness generalization, \eg, overfitting to the target dataset.
Previous methods such as linear interpolation \cite{wortsman2022robust} and parameter regularization \cite{lee2019mixout} mitigate overfitting by introducing constraints in the parameter space.
Although these methods can apply to adversarial fine-tuning and alleviate overfitting, they do not focus on improving the model's zero-shot robust accuracy as their optimization goal, thereby being limited compared with adding constraints in the feature space.
%在参数空间施加的约束

Inspired by the impressive zero-shot performance of the foundational model (\eg, CLIP), we introduce the Pre-trained Model Guided Adversarial Fine-Tuning (PMG-AFT), a novel method enhancing both the generalizability and adversarial robustness of models.
% When adversarially fine-tuning CLIP, PMG-AFT incorporate extra constraints from the pre-trained model and clean examples into the objective function, encouraging the target model to keep the generalized information learned during pre-training.
Besides the cross-entropy loss between adversarial examples and true labels used in original adversarial training \cite{madry2017towards}, PMG-AFT incorporates additional constraints from the pre-trained model and clean examples into the objective function, encouraging the target model to retain the generalized information learned during pre-training.
Specifically, PMG-AFT conducts an auxiliary generalization information branch, which minimizes the distance between the adversarial example outputs in the target model and the pre-trained model.
A regularization loss is also introduced to further enhance the model's adversarial robustness generalization capabilities.
We conduct extensive experiments on additional 15 zero-shot datasets to evaluate our method.
Our method outperforms the state-of-art methods with a significant improvement of up to 4.99\% in terms of average robust accuracy. 
Moreover, attributed to the effectiveness of the generalization information branch, our method also achieves an 8.72\% increase in average accuracy on clean samples, demonstrating the superiority of our PMG-AFT approach again.

Our main contributions are summarized as follows:
\begin{itemize}
    \item We propose the PMG-AFT method, which learns generalization features from the original pre-trained model, effectively mitigating overfitting and enhancing the zero-shot adversarial robustness of the CLIP model.
    \item Our method introduces improvements during the parameters update phase of adversarial fine-tuning and can effectively combine with the defense framework that originates from the perspective of adversarial example generation.
    \item Extensive experiments demonstrate that PMG-AFT consistently outperforms the state-of-the-art in terms of zero-shot robust accuracy and clean accuracy.
\end{itemize}

\section{Related Work}

\textbf{Pre-trained Vision-language Model.}
In recent years, the fusion of vision and language understanding has garnered widespread attention. 
% Early research in this multi-modal field primarily focused on task-specific models, such as image captioning \cite{vinyals2015show} or visual question answering \cite{antol2015vqa,malinowski2014multi}. 
% However, the emergence of pre-trained vision-language models has profoundly transformed this domain.
Inspired by the success of pre-trained language models like BERT \cite{devlin2018bert} and GPT \cite{chen2022visualgpt}, the rise of Self-Supervised Learning (SSL) \cite{balestriero2023cookbook} in computer vision has started to produce task-agnostic models with foundational characteristics, which we refer to as "foundational models."
For instance, models such as SimCLR \cite{chen2020simple}, VL-bert \cite{su2019vl}, DINO \cite{caron2021emerging}, and DINOv2 \cite{oquab2023dinov2} learn representations from unlabeled images and have demonstrated impressive flexibility in addressing various downstream tasks.
Furthermore, models like CLIP \cite{radford2021learning}, SWAG \cite{singh2022revisiting}, ALBEF \cite{li2022align}, BLIP \cite{li2022blip} incorporate a vision-language component during pretraining, further enhancing downstream adaptability via zero-shot learning.
Contrary to earlier works that aimed at improving performance on standard benchmarks, our study takes a different direction by focusing on the adversarial robustness problem of such pre-trained frameworks and improving their zero-shot adversarial robustness.

\noindent\textbf{Adversarial Robustness.}
Deep neural networks have been found vulnerable to adversarial attacks \cite{goodfellow2014explaining,madry2017towards,dong2018boosting,carlini2017towards,athalye2018obfuscated}, which delicately generate imperceptible noises added to the original images leading to model misclassifications.
To enhance the robustness of neural networks against adversarial examples, a series of defense strategies have been proposed.
Among them, adversarial training \cite{madry2017towards,zhang2019theoretically,wu2020adversarial,shafahi2019adversarial} is considered the most effective defense mechanism, which employs adversarial examples during the training phase and integrates them into the training dataset. 
It notably enhances the robustness of deep neural networks against adversarial attacks.
With the rise of large-scale pre-trained vision-language (VLP) model, the robustness of VLP model has gradually come under investigation, and numerous attack algorithms targeting them have emerged \cite{zhao2023evaluating,zhang2022towards,inkawhich2023adversarial,lu2023set}.
In our study, we explore the performance of adversarial training on VLP models, aiming to enhance their adversarial robustness for various downstream tasks.
Mao et al. \cite{mao2022understanding} investigates this issue from the perspective of adversarial example generation and proposes an adversarial fine-tuning algorithm supervised by text.
Li et al. \cite{li2023anchor} adjusts the embedding of the textual modality, reducing the similarity between labels, thus enhancing robustness. 
In contrast to their approaches, we tackle the issue from the perspective of the training process itself, encouraging the adversarial fine-tuned model to leverage the generalized features from the original model for improving the adversarial robustness.

\noindent\textbf{Finetuning and Catastrophic Overfitting.}
Fine-tuning is a prevalent strategy aimed at adapting pre-trained models to specific downstream tasks \cite{devlin2018bert,dosovitskiy2020image}. 
However, when fine-tuning pre-trained vision-language models, one may encounter overfitting issues. 
During this process, the target model might deviate significantly from the pre-trained one, leading to overfitting on small-scale fine-tuning datasets \cite{dodge2020fine,zhang2020revisiting}.
Within the adversarial training framework, the issue of overfitting still exists and is referred to as "adversarial overfitting" \cite{rice2020overfitting,wang2023balance}. 
In this situation, the model is influenced by the distribution of adversarial examples, leading to a significant decline in both robust accuracy and clean accuracy.
Methods to address overfitting include linear interpolation \cite{wortsman2022robust}, and parameter regularization \cite{lee2019mixout}. 
These methods focus on parameter space to constrain the distance between two models while our method addresses overfitting by encouraging memory in the feature space.

\section{Methodology}

\hspace*{0.422cm}We first give background in Sec. \ref{sec:Preliminaries} on adversarial attacks, adversarial training, and zero-shot adversarial robustness.
In Sec. \ref{sec:problem analysis}, we discuss that adversarial fine-tuning can lead to model overfitting, damaging the generalization of the target model.
Finally, Sec. \ref{sec:method detail} provides a detailed introduction to our Pre-trained Model Guided Adversarial Fine-Tuning (PMG-AFT) method, including its various components and the loss function.

\subsection{Preliminaries and Problem Setup}
\label{sec:Preliminaries}
\hspace*{0.422cm}In this paper, we select the image classification task to introduce our method, yet it can be also extended to other tasks.
For large-scale pre-trained vision-language (VLP) models, we choose the CLIP model, one of the typical VLP models for zero-shot recognition, as our base model.
Let $F_{\theta}(\cdot)$ represent the CLIP image encoder parameterized by $\theta$ and $T_{\phi}(\cdot)$ represent the CLIP text encoder parameterized by $\phi$.
Given an input image $x$ and a textual description about its category, such as "This is a photo of a \{\}", denoted as $t$, the model will provide an image representation $F_{\theta}(x)$ and a text representation $T_{\phi}(t)$.
For image classification, we compute the similarity between $F_{\theta}(x)$ and each candidate category embedding $T_{\phi}(t_i)$, where $t_i$ represents a certain prompt of one category, to obtain a $c$-dimensional output, where $c$ represents the number of categories, and the category with the highest similarity is selected as the classification result.
For training or fine-tuning, the model updates its parameters by minimizing the cross-entropy loss, $L([\text{sim}(F_{\theta}(x), T_{\phi}(t_1)), \dots, \text{sim}(F_{\theta}(x), T_{\phi}(t_c))],y)$, where $y$ is the one-hot vector label.
For convenience of representation, we denote it as $L(x,t,y)$.

\noindent\textbf{Adversarial Attacks.} Adversarial attacks typically refer to adding an imperceptible, optimizable perturbation to a clean image, misleading the model to produce an incorrect prediction.
A well-known white-box attack method is Projected Gradient Descent (PGD) \cite{madry2017towards}, which uses multi-step gradient ascent steps to maximize the cross entropy loss while projecting intermediate perturbation to the specified region constrained by the p-norm:
\begin{equation}
\begin{aligned}
\label{pgd}
x_{k+1} &= x_k + \alpha \nabla_{x_k} L(x_k,t,y), \\
x_{k+1} &= P_{B(x,\varepsilon)}(x_{k+1}), \quad k = 0, \dots, K-1,
\end{aligned}
\end{equation}
where $x_0 = x+\varepsilon_0$, $\varepsilon_0 \sim U[0, \varepsilon]$, $B(x,\varepsilon)=\{x|\ ||x-x_k||_p<\varepsilon\}$, $\varepsilon$ represents the perturbation bound under p-norm .
In the subsequent descriptions of the paper, we denote the adversarial examples as $x_a$.

\noindent\textbf{Adversarial Fine-Tuning.} Adversarial training formulates the training process into a min-max problem and can be described as:
\begin{equation}
\begin{aligned}
\label{minmax game}
\min_\theta \mathbb{E}_{x, y \sim P_D} \left[ \max_{x_a \in B(x,\varepsilon)} L(x_a,t,y) \right].
\end{aligned}
\end{equation}
The inner maximization problem aims to find a stronger adversarial example, while the outer minimization problem optimizes the model parameters so that the model $F_{\theta}$ still makes correct predictions on the adversarial examples.
It should be noted that here we only adjust the parameters of the image encoder $\theta$, while the parameters of the text encoder $\phi$ remain unchanged.
Since fine-tuning is generally used to adapt CLIP to downstream tasks, such an objective function \eqref{minmax game} can also be applied to the fine-tuning of CLIP towards robustness, and we refer to it as adversarial fine-tuning.

\noindent\textbf{Zero-Shot Adversarial Robustness.} 
Whether the excellent generalization capabilities of large-scale visual-language models on new tasks and datasets can remain consistent in the field of adversarial defense is a question worth investigating.
We use zero-shot robust accuracy to measure this type of generalization.
During the training phase, we select a target dataset such as TinyImageNet \cite{deng2009imagenet} and perform adversarial fine-tuning on the model.
During the testing phase, we generate adversarial examples using white-box attacks to test CLIP, evaluating the zero-shot accuracy of these adversarial examples across multiple datasets.

\subsection{Adversarial Fine-Tuning is Prone to Overfitting}
\label{sec:problem analysis}
\hspace*{0.422cm}It is well-known that pre-trained models tend to overfit during fine-tuning, especially when the dataset for the downstream task is small \cite{dodge2020fine}. 
During the fine-tuning process, a target model continuously deviates from the pre-trained model to adapt to the downstream task.
From Fig. \ref{fig:fig1b}, we can easily observe that the model fine-tuned on clean TinyImageNet, compared with the original CLIP, only shows improved accuracy when tested on TinyImageNet itself, while experiencing a decline on other zero-shot datasets.
This demonstrates the presence of overfitting.
% In recent years, the notion of foundational models has gained increasing attention. 
% Models are expected to have the capacity to tackle a wide range of tasks. Indeed, fine-tuning boosts performance on specific downstream tasks, but it compromises the capabilities of the foundational model.

\begin{figure}[t]
    \centering
    \includegraphics[scale=0.2]{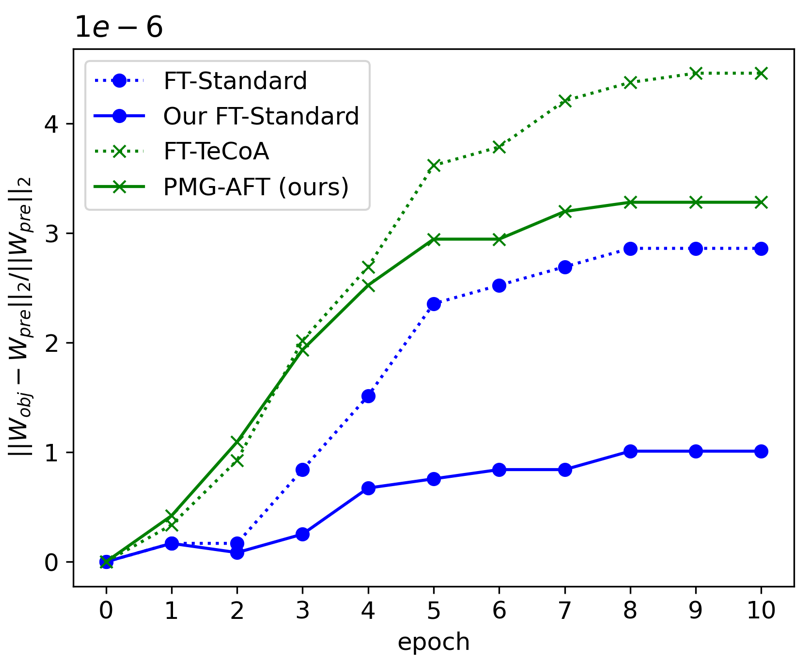}
    \vspace{-3mm}
    \caption{Relative $L_2$ distance between CLIPs fine-tuned on TinyImageNet using various strategies and original CLIP model in the parameter space. Our FT-Standard represents the application of our proposed fine-tuning method on clean target datasets.}
    \label{fig:figure2}
    \vspace{-5mm}
\end{figure}

Adversarial fine-tuning aims to leverage adversarial examples to enhance the robustness of the model. 
The distribution of these adversarial examples often lies far beyond the manifold of the clean example distribution. 
To gain the capability of correctly classifying these adversarial examples, the target model is fine-tuned towards a solution that deviates even more from its initial optimization point, namely the pre-trained model.
Fig.\ref{fig:figure1} also illustrates this point. 
The CLIP model, after fine-tuned by FT-TeCoA \cite{mao2022understanding}, has indeed improved the adversarial robustness across various datasets. 
However, its accuracy on clean datasets has significantly decreased.
This indicates that adversarial fine-tuning also experiences the issue of overfitting.
To more intuitively demonstrate the overfitting phenomenon, we examined the changes in parameters during the training process, as shown in Fig. \ref{fig:figure2}.
As training progresses, the parameters of the standard fine-tuning and adversarial fine-tuning CLIP model both deviate from the original CLIP model, with the adversarial fine-tuning deviating to a greater extent.
This overfitting phenomenon directly leads to a decrease in the model's generalization capability.

\begin{figure*}
\centering
\includegraphics[width=0.9\textwidth]{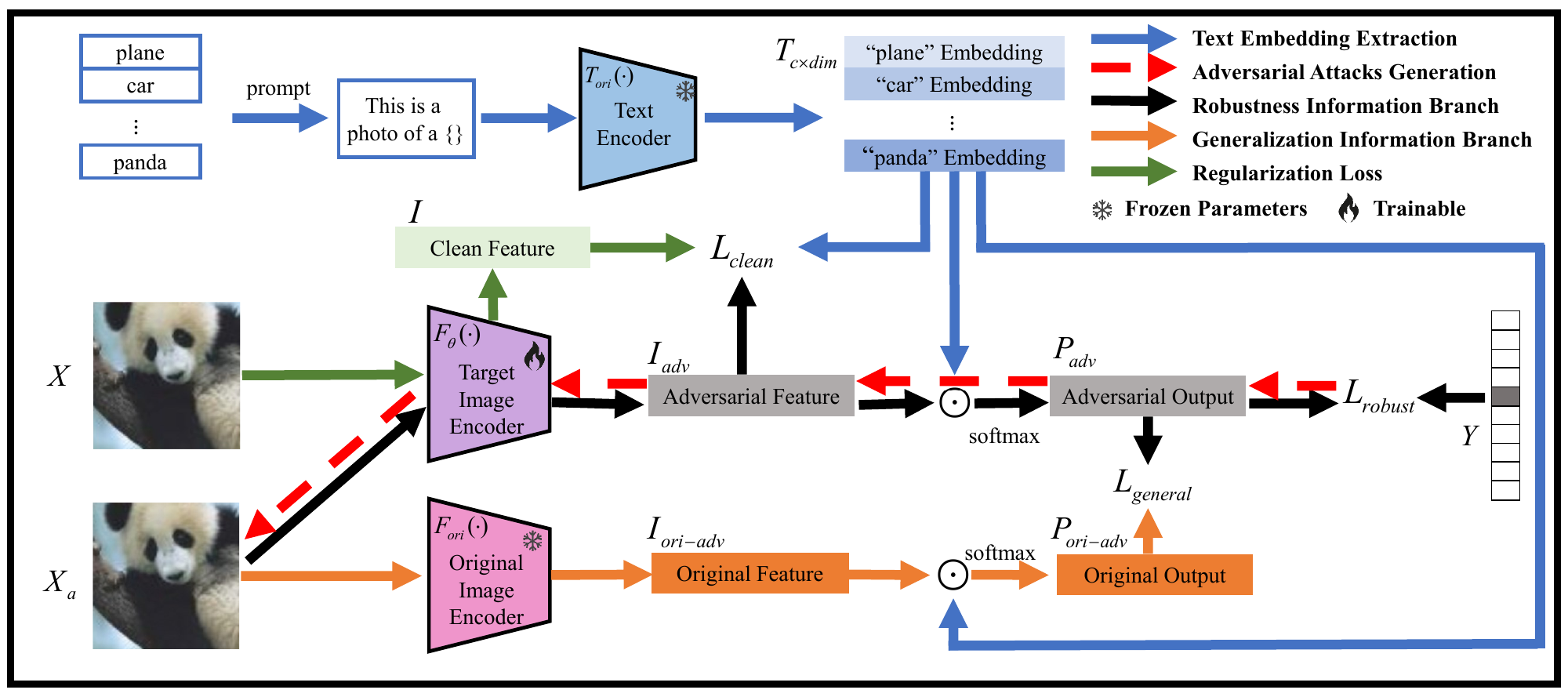}
\caption{The pipeline of PMG-AFT. PMG-AFT first uses the text encoder from the pre-trained CLIP model to obtain text embeddings, then employs TeCoA\cite{mao2022understanding} loss which is $L_{robust}$ in our method to generate adversarial examples. During the model parameter update phase, we split into two branches: the robustness information branch, which maximizes the similarity between the output of the target model and the GT via $L_{robust}$, generalization information branch maximizes the output of the adversarial samples between the target model and the original model via $L_{general}$. A regularization loss ($L_{clean}$) is applied to the adversarial and clean outputs. Only the image encoder of the target model can be trained and the adversarial examples generation alternates with parameters updating. $\odot$ means matrix inner product.}
\label{fig:pipeline}
\vspace{-5mm}
\end{figure*}

\subsection{PMG-AFT}
\label{sec:method detail}
\hspace*{0.422cm}To mitigate overfitting and enhance generalization, adversarial fine-tuning should retain generalizable features that the pre-trained model has already captured. 
For example, adversarial fine-tuning favors the target model extracting features invariant to perturbations for robustness. 
Therefore, all the generalizable and robust information captured during pre-training are especially valuable and should be memorized.
To achieve this objective, we propose PMG-AFT.
PMG-AFT first obtains text embeddings from the frozen CLIP model's text encoder, then uses these text embeddings to supervise the generation of adversarial examples for adversarial fine-tuning. 
The entire fine-tuning process is divided into two branches: the robustness information branch aims to enhance the model's robustness for a specific task, while the generalization information branch aims to retain the generalization capabilities of the original CLIP model. 
The pipeline of PMG-AFT is shown in Fig. \ref{fig:pipeline}, and we will describe the design of PMG-AFT in detail.

\noindent\textbf{Text Embedding Extraction.}
As mentioned in Sec. \ref{sec:Preliminaries}, we use the text encoder from the pre-trained CLIP model, denoted as $T_{ori}(\cdot)$ to extract text embeddings of the category prompts we designed, which are used for the generation of adversarial examples and supervision information for adversarial fine-tuning.
We organize the text embeddings into the form of a matrix, denoted as $T \in \mathbb{R}^{c \times dim}$, where $c$ is the number of categories, $dim$ is the dimension of embedding, each row represents the embedding of a category. 
It's worth noting that our primary focus is on attacks in the image modality. 
Hence, the text modality encoder is frozen and does not participate in the adversarial fine-tuning parameter updates. 

\noindent\textbf{Adversarial Attacks Generation.}
Adversarial fine-tuning requires the adversarial examples generated from the training dataset. 
In the phase of generating adversarial examples, we use the PGD \cite{madry2017towards} method described in Equ. \eqref{pgd}.
The specific calculation process is as follows. We pre-set the attack step size $\alpha$, the number of iterations $K$, and the perturbation range $\varepsilon$.
We denote the image encoder of the target model as $F_{\theta}(\cdot)$, the embeddings of a batch of natural images $X$ as $I \in \mathbb{R}^{N \times dim}$, where $N$ is the size of a batch, $dim$ is the dimension of embedding, each row represents the embedding of an input image.
Based on the inference process of CLIP, we calculate the similarity matrix through the matrix inner product:
\vspace{-1mm}
\begin{equation}
\begin{aligned}
S = I \cdot T^\top.
\end{aligned}
\end{equation}
Next, to transform these similarity scores into a probability distribution, we apply the softmax function:
\begin{equation}
\begin{aligned}
P_{ij} = \frac{\exp(S_{ij})}{\sum_{k=1}^c \exp(S_{ik})},
\end{aligned}
\end{equation}
where $P_{ij}$ is the predicted probability of the $i$-$th$ image belonging to the $j$-$th$ class.
Finally, given the one-hot ground truth label $Y$ (with a size of $N\times c$) of a batch, the cross-entropy loss can be represented as:
\begin{equation}
\label{mingame}
\begin{aligned}
L_{CE}(X,T,Y) = -\frac{1}{N} \sum_{i=1}^{N} \sum_{j=1}^c Y_{ij} \log(P_{ij}).
\end{aligned}
\end{equation}
By incorporating the loss function and the aforementioned hyperparameters into Equ. \eqref{pgd}, we can obtain the adversarial examples $X_a$.

\noindent\textbf{Robustness Information Branch.}
To enhance the robustness of the target model, we need to update the model parameters through the outer minimization problem in Equ. \eqref{minmax game}. 
Its goal is to minimize the model's loss on adversarial examples. 
The most intuitive and effective loss is the classification cross-entropy loss on the adversarial examples, encouraging the target model to correctly classify adversarial examples to achieve robustness.
We take the adversarial examples $X_a$ as input and minimize the loss function in Equ. \eqref{mingame} to optimize the parameters $\theta$ of the image encoder $F_{\theta}(\cdot)$,  obtaining one term of the loss function,
\begin{equation}
\begin{aligned}
L_{robust} = L_{CE}(X_a,T,Y). 
\end{aligned}
\end{equation}
It encourages the similarity of the adversarial visual feature to be as close as possible to the textual representation of the true class label. 
% Under the supervision of CE loss, the output feature can be optimized to be far away from all other categories.

\noindent\textbf{Generalization Information Branch.}
As adversarial examples are generated based on a specific dataset, the robustness features acquired by the target model are confined and may excessively specialize to a particular downstream task dataset, consequently compromising generalizability.
Our method introduces a generalization information branch to improve the generalization of adversarial robustness.
The main body of this branch is an image encoder of the original pre-trained CLIP, which is denoted as $F_{ori}(\cdot)$.
$I_{ori-adv}$ denotes the embeddings of adversarial images it produces. 
Since the pre-trained model is a fixed deterministic function, the supervision objective of this branch essentially encourages the target model 
$F_{\theta}(\cdot)$ to output features that can predict the information of the original foundational model as much as possible, thereby helping to mitigate the problem of overfitting.
We feed the adversarial examples $X_a$ into both the target model and the original pre-trained model, obtaining both prediction results based on the text embeddings $T$. 
\begin{equation}
\begin{aligned}
P_{adv} &= \text{softmax}(I_{adv} \cdot T^\top),\\
P_{ori-adv} &= \text{softmax}(I_{ori-adv} \cdot T^\top).
\end{aligned}
\end{equation}
We measure the distance between them using the KL divergence, aiming to minimize this distance as much as possible in order to learn knowledge from the original model.
\begin{equation}
\begin{aligned}
D_{KL}(P || Q) = \sum_{i} P(i) \log\left(\frac{P(i)}{Q(i)}\right), 
\end{aligned}
\end{equation}
\vspace{-2mm}
\begin{equation}
\begin{aligned}
L_{general} = \frac{1}{N} \sum_{j=1}^{N} D_{KL}(P_{adv_j}|| P_{ori-adv_j}),
\end{aligned}
\end{equation}
where $P(i)$ represents the $i$-$th$ element in the output probability tensor $P$, and $j$ represents the $j$-$th$ sample in a batch.

\noindent\textbf{Loss Function.}
In addition to the two terms of loss mentioned above, we further introduce a regularization loss.
Specifically, it encourages the features of adversarial images to be similar to those of clean images in the target model:
\begin{equation}
\begin{aligned}
P_{clean} &= \text{softmax}(I \cdot T^\top),\\
L_{clean} &= \frac{1}{N} \sum_{j=1}^{N} D_{KL}(P_{adv_j}|| P_{clean_j}).
\end{aligned}
\end{equation}
The regularization loss is independent of the labels of the training examples and helps to maintain the generalization ability of the original CLIP model.
Combined with generalization information branch, the regularization loss can further enhance the model's adversarial robustness on unseen categories. See Sec. \ref{lossterm} for detailed results.

In summary, the loss function during the training process is $L=L_{robust}+\alpha L_{general}+\beta L_{clean}$, where $\alpha$ and $\beta$ are hyper-parameters.

Our algorithm iteratively alternates between generating adversarial examples and updating the model parameters. 
As PMG-AFT utilizes additional information from the original model to correct the fine-tuning process of the target model, it helps the model maintain adversarial robustness in terms of zero-shot generalization capability.
%可以考虑删除
\section{Experiments}
\subsection{Experimental Setup}

\noindent\textbf{Datasets.} We fine-tune the CLIP model on the TinyImageNet \cite{deng2009imagenet} dataset, and conduct evaluations on TinyImageNet \cite{deng2009imagenet} as well as additional 15 zero-shot datasets, reporting their robust accuracy and clean accuracy.
Specifically, these 15 datasets fall into 5 categories: general object recognition including CIFAR10 \cite{krizhevsky2009learning}, CIFAR100 \cite{krizhevsky2009learning}, STL10 \cite{coates2011analysis}, ImageNet \cite{deng2009imagenet}, Caltech101 \cite{fei2006one}, and Caltech256 \cite{griffin2007caltech}; fine-grained recognition such as OxfordPets \cite{parkhi2012cats}, Flowers102 \cite{nilsback2008automated}, FGVCAircraft \cite{maji2013fine}, and StanfordCars \cite{krause20133d}; scene recognition represented by SUN397 \cite{xiao2010sun}; domain-specific data which includes Food101 \cite{bossard2014food}, EuroSAT \cite{helber2019eurosat}, and DTD \cite{cimpoi2014describing}; medical image, which in this case is PCAM \cite{bejnordi2017diagnostic}.
We also conduct experiments with the same settings on the CIFAR100 dataset and evaluations on the same 16 datasets.
To accommodate the input requirements of the CLIP model, they have all been preprocessed to a size of $3 \times 224 \times 224$.

\noindent\textbf{Baseline.} Considering that research on zero-shot adversarial robustness is in its nascent stages with a limited number of available methods, our primary comparisons are focused on evaluating our approach against the current SOTA method named FT-TeCoA \cite{mao2022understanding}.
% TeCoA \cite{mao2022understanding} enhances the zero-shot robust accuracy of the CLIP model by generating adversarial examples supervised by text modality. 
Following FT-TeCoA \cite{mao2022understanding}, we also employ two adaptation methods, \ie, fine-tuning and visual prompt for comprehensive evaluations.

\begin{table*}[htbp]
\centering
\caption{Adversarial zero-shot robust accuracies under PGD-10 \cite{madry2017towards} attack. We fine-tune the model on TinyImageNet \cite{deng2009imagenet} and evaluate six methods (rows) on 16 datasets (columns), presenting the accuracy for each dataset as well as the average accuracy, with the best results shown in bold. 
The accuracy here is percentage.}
\label{tab:tiny-robust}
\begin{adjustbox}{width=\textwidth,keepaspectratio}
\begin{tabular}{lcccccccccccccccccc}
\toprule
% & \multicolumn{16}{c}{\textbf{Datasets}} \\ % 合并中间的列，并在合并的列上添加标题 "Datasets"
% \cmidrule(lr){2-17} % 这个命令将会在从第二列到第十三列的上方画一条线，'lr'是为了减少线的长度
Method    & \footnotesize\rotatebox{45}{CIFAR10} & \footnotesize\rotatebox{45}{CIFAR100} & \footnotesize\rotatebox{45}{STL10} & \footnotesize\rotatebox{45}{SUN397} & \footnotesize\rotatebox{45}{Food101} & \footnotesize\rotatebox{45}{Oxfordpet} & \footnotesize\rotatebox{45}{Flowers102} & \footnotesize\rotatebox{45}{DTD}   & \footnotesize\rotatebox{45}{EuroSAT} & \footnotesize\rotatebox{45}{Fgvc\_Aircraft} & \footnotesize\rotatebox{45}{TinyImageNet} & \footnotesize\rotatebox{45}{ImageNet} & \footnotesize\rotatebox{45}{Caltech101} & \footnotesize\rotatebox{45}{Caltech256} & \footnotesize\rotatebox{45}{StanfordCars} & \footnotesize\rotatebox{45}{PCAM}  & \footnotesize\rotatebox{45}{Average} & \footnotesize\rotatebox{45}{Time (s)} \\ 
\midrule
CLIP           & 42.18 & 17.57 & 67.77 & 10.49 & 3.28 & 8.98 & 4.88 & 18.75 & 3.84 & 0.39 & 0.39 & 6.09 & 15.82 & 23.76 & 2.34 & 51.61 &17.38 & 0    \\
FT-Standard    & 24.21 & 12.30 & 47.65 & 5.71  & 2.65 & 1.56 & 7.61 & 17.57 & 0.13 & 0.00 & 0.39 & 6.87 & 20.31 & 23.82 & 1.75 & 48.15 &13.79 & 234  \\
FT-TeCoA \cite{mao2022understanding}       & 40.82 & 24.41 & 70.70 & 19.21 & 14.45& 28.13& 23.05& 28.13 & 12.57& 3.13 & \textbf{19.33}& 16.48& 24.02 & 40.56 & 12.69& 53.68 &26.96 & 551  \\
PMG-AFT (ours)         & \textbf{66.99} & \textbf{38.28} & \textbf{76.17} & \textbf{24.23} & \textbf{14.92}& \textbf{33.59}& \textbf{23.43}& \textbf{34.38} & \textbf{24.15}& \textbf{3.91} & 14.84& \textbf{17.26}& \textbf{24.02} & \textbf{45.05} & \textbf{14.64}& \textbf{57.64} &\textbf{31.95} & 849  \\ \hline
VP-TeCoA  \cite{mao2022understanding}     & 36.71 & 16.21 & 57.61 & 6.05  & 4.45 & 13.67& 13.02& \textbf{10.15} & 9.73 & 0.00 & 18.55& 5.56 & 18.16 & 25.32 & 2.14 & 54.92 &18.27 & 364  \\
VPT-PMG-AFT (ours)         & \textbf{50.19} & \textbf{22.07} & \textbf{64.45} & \textbf{12.79} & \textbf{8.35 }& \textbf{18.55}& \textbf{16.85}& 3.45  & \textbf{11.08}& \textbf{2.73} & \textbf{20.70}& \textbf{9.92} & \textbf{19.25} & \textbf{38.99} & \textbf{5.46} & \textbf{58.71} &\textbf{22.74} & 661  \\
\bottomrule
\end{tabular}
\end{adjustbox}
\vspace{-4mm}
\end{table*}
\begin{table*}[htbp]
\centering
\caption{Zero-shot clean accuracies. 
After fine-tuning or adversarial fine-tuning on TinyImageNet \cite{deng2009imagenet}, the zero-shot accuracy of the CLIP model on clean images generally decreases. 
Compared with other fine-tuning methods, our approach achieves the best clean accuracy.}
\label{tab:tiny-clean}
\begin{adjustbox}{width=\textwidth,keepaspectratio}
\begin{tabular}{lcccccccccccccccccc}
\toprule
% & \multicolumn{16}{c}{\textbf{Datasets}} \\ % 合并中间的列，并在合并的列上添加标题 "Datasets"
% \cmidrule(lr){2-17} % 这个命令将会在从第二列到第十三列的上方画一条线，'lr'是为了减少线的长度
Method    & \footnotesize\rotatebox{45}{CIFAR10} & \footnotesize\rotatebox{45}{CIFAR100} & \footnotesize\rotatebox{45}{STL10} & \footnotesize\rotatebox{45}{SUN397} & \footnotesize\rotatebox{45}{Food101} & \footnotesize\rotatebox{45}{Oxfordpet} & \footnotesize\rotatebox{45}{Flowers102} & \footnotesize\rotatebox{45}{DTD}   & \footnotesize\rotatebox{45}{EuroSAT} & \footnotesize\rotatebox{45}{Fgvc\_Aircraft} & \footnotesize\rotatebox{45}{TinyImageNet} & \footnotesize\rotatebox{45}{ImageNet} & \footnotesize\rotatebox{45}{Caltech101} & \footnotesize\rotatebox{45}{Caltech256} & \footnotesize\rotatebox{45}{StanfordCars} & \footnotesize\rotatebox{45}{PCAM}  & \footnotesize\rotatebox{45}{Average} & \footnotesize\rotatebox{45}{Time (s)} \\
\midrule
CLIP          & 89.52  & 62.50  & 97.46  & 55.25  & 75.00  & 85.93  & 48.82  & 41.79  & 50.84  & 8.70   & 56.03   & 61.99   & 32.42   & 84.76   & 55.27
   & 53.40  &59.98  &0\\ \hline
FT-Standard   & 88.67  & 58.78  & 95.89  & 46.61  & 70.07  & 75.00  & 41.60  & 41.79  & 38.60  & 8.57   & 58.59   & 51.28   & 29.10   & 79.88   & 37.10   & 54.74   &54.76  &234\\
FT-TeCoA \cite{mao2022understanding}      & 66.79  & 41.01  & 89.25  & 47.01  & 52.81  & 70.31  & 36.13  & 35.94  & 18.88  & 7.81   & 48.83   & 43.67   & 28.32   & 72.98   & 37.89   & 37.89   &46.99 &551\\
PMG-AFT (ours)         & 83.98  & 58.39  & 92.97  & 56.41  & 66.40  & 84.76  & 42.96  & 41.02  & 35.28  & 6.25   & 46.87   & 56.75   & 30.46   & 82.94   & 48.24   & 48.24   &\textbf{55.71} &849\\ \hline
VPT-TeCoA \cite{mao2022understanding}    & 53.51  & 26.36  & 74.41  & 17.22  & 15.85  & 33.20  & 22.39  & 12.50  & 13.77  & 0.39   & 44.92   & 15.82   & 21.26   & 42.38   & 3.32    & 58.71    &28.50 &364\\
VPT-PMG-AFT (ours)       & 67.38  & 33.78  & 81.64  & 29.19  & 28.51  & 49.60  & 17.18  & 13.92  & 13.92  & 7.81   & 52.14   & 26.83   & 21.31   & 61.19   & 16.99   & 58.95   &\textbf{36.03} &661\\
\bottomrule
\end{tabular}
\end{adjustbox}
\vspace{-2mm}
\end{table*}
\begin{table*}[htbp]
\centering
\caption{Zero-shot robust accuracies under AutoAttack \cite{croce2020reliable} with $\varepsilon=1/255$. The dataset used for fine-tuning remains TinyImageNet \cite{deng2009imagenet}.}
\label{tab:AA}
\begin{adjustbox}{width=\textwidth,keepaspectratio}
\begin{tabular}{lcccccccccccccccccc}
\toprule
% & \multicolumn{16}{c}{\textbf{Datasets}} \\ % 合并中间的列，并在合并的列上添加标题 "Datasets"
% \cmidrule(lr){2-17} % 这个命令将会在从第二列到第十三列的上方画一条线，'lr'是为了减少线的长度
Method    & \footnotesize\rotatebox{45}{CIFAR10} & \footnotesize\rotatebox{45}{CIFAR100} & \footnotesize\rotatebox{45}{STL10} & \footnotesize\rotatebox{45}{SUN397} & \footnotesize\rotatebox{45}{Food101} & \footnotesize\rotatebox{45}{Oxfordpet} & \footnotesize\rotatebox{45}{Flowers102} & \footnotesize\rotatebox{45}{DTD}   & \footnotesize\rotatebox{45}{EuroSAT} & \footnotesize\rotatebox{45}{Fgvc\_Aircraft} & \footnotesize\rotatebox{45}{TinyImageNet} & \footnotesize\rotatebox{45}{ImageNet} & \footnotesize\rotatebox{45}{Caltech101} & \footnotesize\rotatebox{45}{Caltech256} & \footnotesize\rotatebox{45}{StanfordCars} & \footnotesize\rotatebox{45}{PCAM}  & \footnotesize\rotatebox{45}{Average} & \footnotesize\rotatebox{45}{Time (s)} \\
\midrule
CLIP     & 8.98  & 3.90  & 13.47 & 0.07  & 0.39  & 0.00  & 0.00  & 0.00 & 0.00   & 0.39  & 0.00 & 0.00  & 0.41 & 0.15 & 0.09  & 0.16 & 1.74 & 0\\
FT-TeCoA \cite{mao2022understanding} & 9.96  & 4.68  & 23.24 & 0.21  & 2.73  & 5.44  & 5.17  & 14.85 & 9.73  & 0.39  & \textbf{16.99} & 8.85  & 16.21 & 27.08 & 4.49  & 15.26 & 10.33 & 551\\
PMG-AFT (ours)   & \textbf{31.05} & \textbf{13.28} & \textbf{44.33} & \textbf{14.47} & \textbf{10.54} & \textbf{12.78} & \textbf{8.75}  & \textbf{23.56} & \textbf{15.13} & \textbf{0.78}  & 13.39 & \textbf{9.07}  & \textbf{16.40} & \textbf{43.16} & \textbf{12.10} & \textbf{17.84} & \textbf{17.91} &849\\
\bottomrule
\end{tabular}
\end{adjustbox}
\vspace{-3mm}
\end{table*}

\noindent\textbf{Implementation Details.}
% 我们使用CLIP模型的ViT-B/32架构作为骨架，使用SGD优化器对目标模型进行10轮微调。对于微调，我们更新图像编码器的所有参数，学习率为5e-5。对于视觉提示，我们将可学习的参数作为提示加在图像层，尺寸与原图相同以及token层，尺寸为100，视觉提示的学习率为40。我们使用扰动大小为1、2、4的扰动分别用于对抗训练以及零样本对抗测试。
We utilize the ViT-B/32 architecture of the CLIP model as the backbone and use the SGD optimizer to adversarially finetune the target model for 10 epochs. 
For fine-tuning, we update all parameters of the image encoder with a learning rate of 5e-5. 
For visual prompt, we introduce learnable parameters as prompts, adding them to both the image layer with the same dimensions as the original image and the token layer with a dimension of 100. 
The learning rate for the visual prompt is set at 40. 
We use $l_\infty$ norm PGD-2 \cite{madry2017towards} and PGD-10 \cite{madry2017towards} attacks with perturbation bounds of $\varepsilon=1/255, 2/255$ and $4/255$ for both adversarial training and evaluation, respectively.
For hyper-parameters, we set $\alpha$ to 1 and $\beta$ to 1.
Each model in this paper is trained on two NVIDIA GeForce RTX 3090 GPUs.

\subsection{Main Result}
%我们首先在TinyImageNet上对模型进行微调，然后在包括其在内的16个数据集上进行评估，这里除了TinyImageNet数据集以外在其他数据集上都是零样本测试，测试时我们使用PGD-10攻击，扰动半径为1.对抗准确率结果如表一所示，干净样本准确率结果如表2所示，每一行代表一个方法在不同测试集上的测试结果。除TinyImageNet数据集外，其他数据集的全部数据(包括训练集)未出现在我们的训练过程中，我们将最好的对抗准确率结果进行了加粗。
\hspace*{0.422cm}We fine-tune the model on TinyImageNet and subsequently evaluate it on all 16 datasets. 
It's noteworthy that, on datasets other than TinyImageNet, the evaluation is conducted in a zero-shot manner.
During training and evaluation, we use the PGD-10 \cite{madry2017towards} attack with a perturbation bound $\varepsilon = 1/255$. 
The robust accuracy results are shown in Tab. \ref{tab:tiny-robust}, and the clean accuracy results are displayed in Tab. \ref{tab:tiny-clean}.
We bold the best robust accuracy results for each dataset.

From Tab. \ref{tab:tiny-robust}, we can see that our method shows an average improvement in robust accuracy of 14.57\% compared with the original CLIP model. 
Compared with FT-TeCoA, the current state-of-the-art, our method shows an average improvement in robust accuracy of 4.99\%, and achieves improvement on the most of datasets except TinyImageNet. 
However, the test on TinyImageNet is not a strict zero-shot test, which also indicates that our method effectively mitigates overfitting on specific downstream tasks. 
Moreover, Tab. \ref{tab:tiny-clean} demonstrates that the improvement in robust accuracy brought by the FT-TeCoA comes at the cost of a 12.99\% decrease in average clean accuracy compared with the original CLIP.
However, our method outperforms FT-TeCoA in terms of clean accuracy, achieving a clean accuracy comparable to the CLIP model, slightly higher than fine-tuning on clean datasets \ie FT-Standard.

For visual prompt, since it involves fewer parameters update than fine-tuning, both robust accuracy and clean accuracy are reduced across different methods. 
Nevertheless, our method consistently outperforms FT-TeCoA, further proving the effectiveness of our approach.
%对于视觉提示，由于涉及的参数更新少于微调，因此无论是对抗准确率还是干净样本准确率都有所降低。尽管如此，我们方法还是一致性的优于TeCoA，再次证明我们方法的有效性。
%因此我们的方法仍然展现出其优越性。
%然而，与微调相比，无论是对抗准确率还是干净样本准确率都有所降低，
%我们也在CIFAR100数据集上进行了相同设定的实验，我们的方法对抗准确率达到了26.06%，相比于FT-TecoA在对抗准确提升了4.71%，具体的实验结果可以见补充材料。

We also conduct experiments with the same setting on the CIFAR100 dataset, where our method achieves an average robust accuracy of 26.06\%, which is an improvement of 4.71\% over FT-TeCoA.
Detailed experimental results can be found in the supplementary materials.
Additionally, since we have introduced an extra branch, the computational overhead is increased compared to FT-TeCoA, resulting in an extra 298 seconds per training epoch. 
However, our PMG-AFT achieves much better results than FT-TeCoA in terms of both robustness and clean accuracies.

\vspace{-1mm}
\subsection{Performance against AutoAttack}
\hspace*{0.422cm}AutoAttack \cite{croce2020reliable}, as a strong attack method, is usually used to further verify the robustness of different models. 
It provides a more comprehensive evaluation of robustness than a single attack algorithm. 
We conduct evaluations using the standard version AutoAttack \cite{croce2020reliable}, and present the robust accuracy at a perturbation bound $\varepsilon$ of $1/255$ in Tab. \ref{tab:AA}.
%在autoattack的攻击下，相比于PGD-10攻击，原始的CLIP模型鲁棒性急剧下降，在各个数据集上平均只有xx的对抗准确率，我们的方法也有一定程度的下降但是仍然提升了模型相比与
Under the AutoAttack, as compared with the PGD-10 \cite{madry2017towards} attack, the robustness of the original CLIP model drops drastically, with an average robust accuracy dropping from 17.38\% to 1.74\% across various datasets. 
Our method also experiences a decline to some extent, but it still enhances the adversarial accuracy of the model compared with both CLIP and FT-TeCoA, demonstrating the effectiveness of our method again.

\begin{figure}[t]
    \centering
    \includegraphics[scale=0.28]{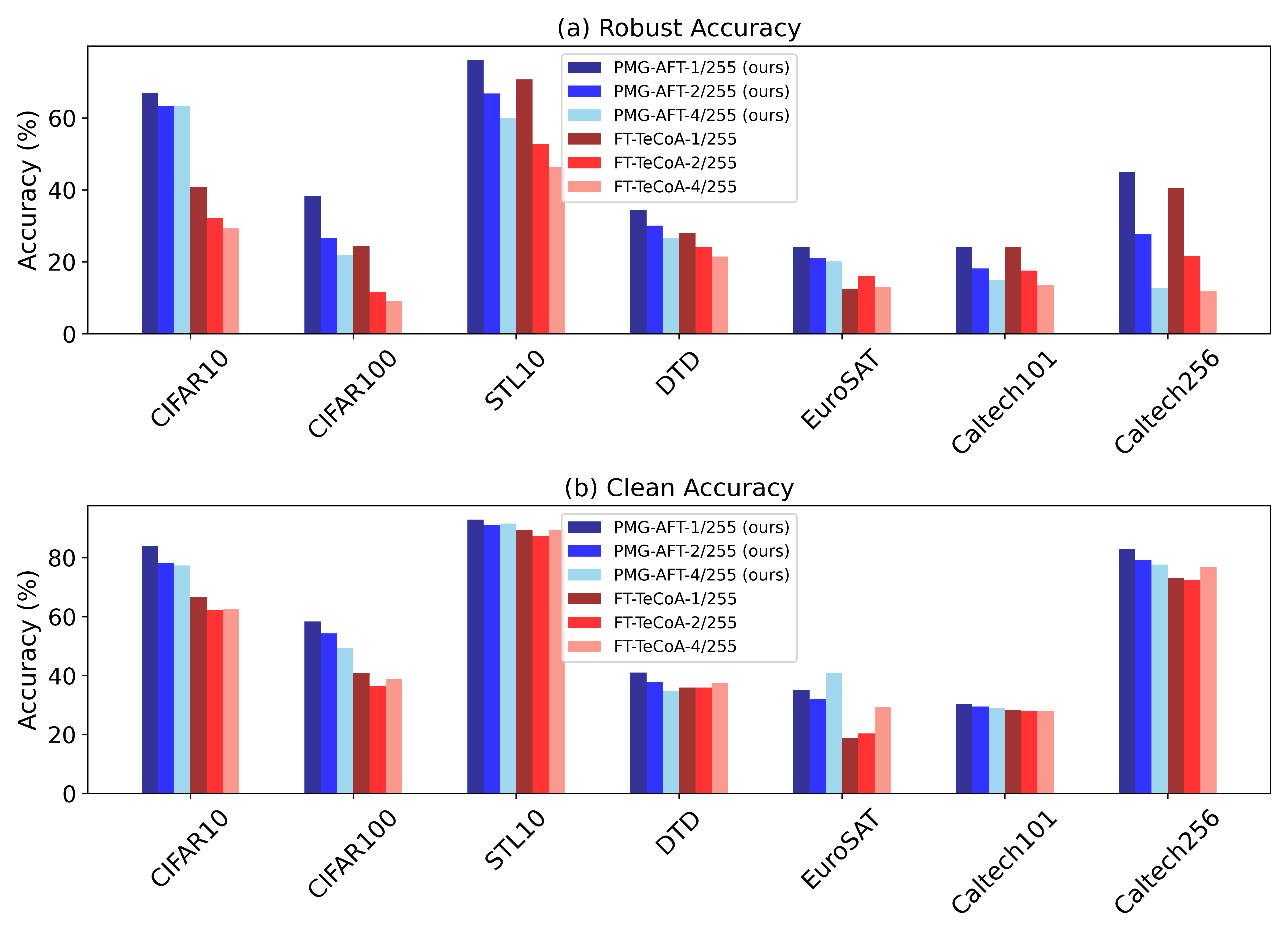}
    \vspace{-4mm}
    \caption{Zero-shot robust accuracy (a) and clean accuracy (b) under different perturbation bounds ($\varepsilon=1/255, 2/255$ and $4/255$). We employ the same perturbation bound for training and testing.}
    \label{fig:strength}
    \vspace{-6mm}
\end{figure}

\vspace{-1mm}
\subsection{Performance against Different Attack Strength}
\hspace*{0.422cm}To verify the impact of changes in adversarial perturbation bounds on our method, we increase the perturbation bound for PMG-AFT from $1/255$ to $4/255$. 
We sampled several datasets to conduct experiments on our method and FT-TeCoA facing different attack bounds. 
The results, as shown in Fig. \ref{fig:strength} indicate that with the increase in adversarial perturbation, both our method and FT-TeCoA experience varying degrees of decline in robust accuracy. 
However, our defense consistently outperforms FT-TeCoA with different attack intensities. 
Besides, as the size of adversarial perturbation continues to increase, the clean accuracy is almost unaffected.
We show detail results on all datasets in the supplementary materials.

\subsection{Trade-off between Robust and Clean Accuracy}
\begin{figure}[t]
    \centering
    \includegraphics[scale=0.43]{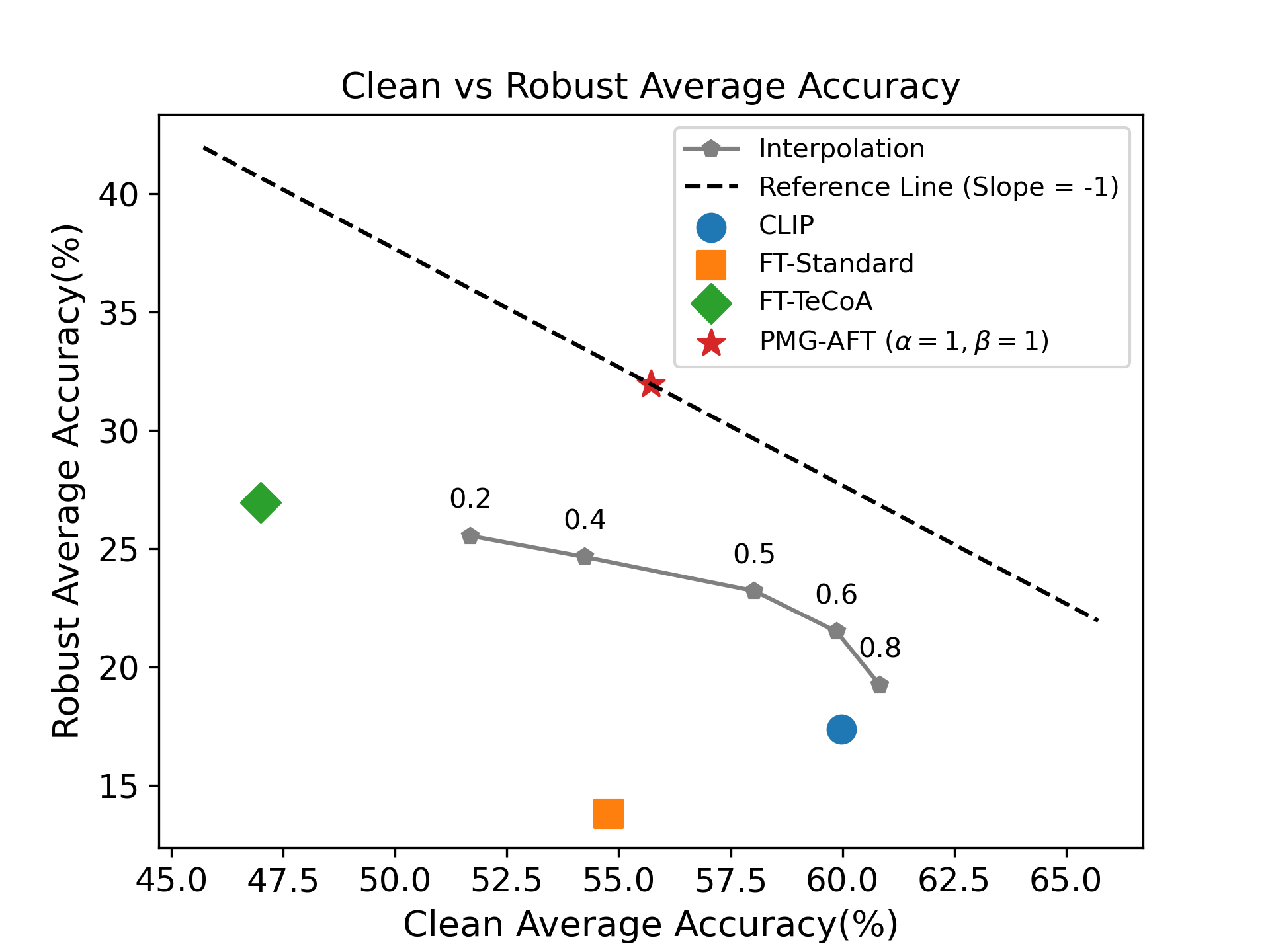}
    \vspace{-2mm}
    \caption{Trade-off between robust and clean accuracy for different fine-tuning methods. Each scatter point represents one method, and the dashed line represents a reference slope line, used to examine the optimal performance of the trade-off.}
    \label{fig:tradeoff}
    \vspace{-5mm}
\end{figure}

\hspace*{0.422cm}Adversarial training often involves a trade-off between clean accuracy and robust accuracy. 
Existing work has approached the trade-off from the perspective of model parameters interpolation \cite{wortsman2022robust}. 
In our experiments, we refer to this method as "interpolation".
We plotted the relationship between the robust accuracy and the clean accuracy of different methods, as shown in Fig. \ref{fig:tradeoff}. 
The $x$-axis represents the average clean accuracy tested on 16 datasets for the original CLIP and various fine-tuned models, while the $y$-axis represents the average robust accuracy. 
It can be observed that our method achieves the most optimal balance among all the methods compared, and it outperforms all other methods in both robust accuracy and clean accuracy.

\subsection{Ablation Study}

\noindent\textbf{Contribution of Loss Function Term.} 
\label{lossterm}
To demonstrate the effectiveness of the auxiliary branch and regularization loss introduced by our method, we incrementally added loss terms and adjusted hyper-parameters. 
The experimental results, as shown in \ref{tab:lossterm}, indicated that after introducing the generalization information branch ($L_{general}$), the model's adversarial robust generalization and clean sample generalization both improved compared with FT-TeCoA, with an average increase of 1.48\% and 11.29\% respectively. 
With the incorporation of the regularization loss, the model's robust generalization experienced a further significant increase, from 28.44\% to 32.04\%, which underscores the effectiveness and considerable potential of the branch we proposed. 
By adjusting different hyper-parameters, we found that the performance of the method is optimal when $\alpha=1$ and $\beta=1$. For detail experimental results, please refer to the supplementary materials.

\begin{table}[htbp]
\centering
\caption{Contribution of the loss function we propose. we incrementally added loss terms and report the average robust accuracy and clean accuracy of the model after fine-tuning.}
\label{tab:lossterm}
\begin{adjustbox}{width=0.45\textwidth,keepaspectratio}
\begin{tabular}{lcc}
\toprule
{Method} & {Avg Clean Acc (\%)} & {Avg Robust Acc (\%)} \\
\midrule
FT-TeCoA \cite{mao2022understanding} ($\alpha=0,\beta=0$)     & 46.99 & 26.96 \\
\hspace*{0.2cm}+$L_{general}$ ($\alpha=1,\beta=0$) (ours)  & \textbf{58.28} & 28.44 \\
\hspace*{0.2cm}+$L_{general}$+$L_{clean}$ ($\alpha=1,\beta=1$) (ours) & 55.72 & \textbf{32.04} \\
\bottomrule
\end{tabular}
\end{adjustbox}
\vspace{-4mm}
\end{table}
\begin{table}[htbp]
\centering
\caption{Average robust and clean accuracy under the selection of different feature layers and distance metric.}
\label{tab:ablation}
\begin{adjustbox}{width=0.4\textwidth,keepaspectratio}
\begin{tabular}{lcc}
\toprule
Method & Avg Clean Acc (\%) & Avg Robust Acc (\%) \\
\midrule
% $\text{Output}_\text{KL}$ & \textbf{55.71} & \textbf{31.95} \\
% $\text{Output}_{\text{L}_2}$ & 48.21 & 27.23 \\
% $\text{Feature}_{\text{L}_2}$ & 38.32 & 22.90 \\
% $\text{Feature}_{\text{COS}}$ & 52.98 & 24.53 \\
Output + KL & \textbf{55.71} & \textbf{31.95} \\
Output + $L_2$ & 48.21 & 27.23 \\
Feature + $L_2$ & 38.32 & 22.90 \\
Feature + COS & 52.98 & 24.53 \\
\bottomrule
\end{tabular}
\end{adjustbox}
\vspace{-2mm}
\end{table}

\noindent\textbf{Impact of Feature Layer and Distance Metric.}
Our PMG-AFT involves minimizing the feature-level distance, as in $L_{general}$.
Therefore, the choice of which feature layer to use and how to measure the distance is crucial to our method. 
As shown in Tab. \ref{tab:ablation}, we compared the results of PMG-AFT when applied to the output layer and penultimate feature layer using KL divergence, $L_2$ distance, and cosine distance. 
Due to space limitations, we only show their average robust and clean accuracies. For the detailed results on each dataset, please refer to the supplementary materials.
From the experimental results, we can see that using KL divergence at the output layer yields the best results in terms of both robust accuracy and clean accuracy.

\vspace{-1mm}
\section{Conclusion}
\hspace*{0.422cm}Inspired by the powerful generalization capabilities of pre-trained models, we propose a novel adversarial fine-tuning method PMG-AFT to enhance the CLIP's zero-shot adversarial robustness.
By incorporating constraints from the pre-trained model and clean examples, generalized features are learned by the target model.
Our method makes improvement from the perspective of the outer minimization problem in adversarial training and can combine with adversarial sample generation algorithms. 
Extensive experiments demonstrate that our method has strong adversarial robustness across multiple zero-shot datasets, and achieves a better trade-off between robust and clean accuracy.

\vspace{-1mm}
\section{Acknowledgement}
\hspace*{0.422cm}This work is partially supported by National Key R\&D Program of China (No. 2021YFC3310100), 
Strategic Priority Research Program of the Chinese Academy of Sciences (No.XDB0680101), 
Beijing Nova Program (20230484368), 
Suzhou Frontier Technology Research Project (No. SYG202325), 
and Youth Innovation Promotion Association CAS.

{
    \small
    \bibliographystyle{ieeenat_fullname}
    \bibliography{main}
}


\end{document}